\documentclass[conference]{IEEEtran}
\IEEEoverridecommandlockouts
\usepackage{cite}
\usepackage{amsmath,amssymb,amsfonts}
\usepackage{algorithmic}
\usepackage{graphicx}
\usepackage{textcomp}
\usepackage{tabularx}
\usepackage{url}
\usepackage{booktabs}
\usepackage{xcolor}
\usepackage{adjustbox}
\usepackage{makecell}
\usepackage{multirow}
\def\BibTeX{{\rm B\kern-.05em{\sc i\kern-.025em b}\kern-.08em
    T\kern-.1667em\lower.7ex\hbox{E}\kern-.125emX}}

\makeatletter
\newcommand{\linebreakand}{%
  \end{@IEEEauthorhalign}
  \hfill\mbox{}\par
  \mbox{}\hfill\begin{@IEEEauthorhalign}
}
\makeatother

\begin{document}

\title{Robust Biomedical Publication Type and Study Design Classification with Knowledge-Guided Perturbations
}

\author{
\IEEEauthorblockN{Shufan Ming}
\IEEEauthorblockA{
\textit{School of Information Sciences}\\
\textit{University of Illinois Urbana-Champaign}\\
Champaign, USA\\
shufanm2@illinois.edu
}
\and
\IEEEauthorblockN{Joe D. Menke}
\IEEEauthorblockA{
\textit{School of Information Sciences}\\
\textit{University of Illinois Urbana-Champaign}\\
Champaign, USA\\
jmenke@illinois.edu
}
\and
\IEEEauthorblockN{Neil R. Smalheiser}
\IEEEauthorblockA{
\textit{Department of Psychiatry}\\
\textit{University of Illinois Chicago}\\
Chicago, USA\\
neils@uic.edu
}
\linebreakand
\IEEEauthorblockN{Halil Kilicoglu}
\IEEEauthorblockA{
\textit{School of Information Sciences}\\
\textit{University of Illinois Urbana-Champaign}\\
Champaign, USA\\
halil@illinois.edu
}
}

\maketitle

\begin{abstract}
Accurately and consistently indexing biomedical literature by publication type and study design is essential for supporting evidence synthesis and knowledge discovery. Prior work on automated publication type and study design indexing has primarily focused on expanding label coverage, enriching feature representations, and improving in-domain accuracy, with evaluation typically conducted on data drawn from the same distribution as training. Although pretrained biomedical language models achieve strong performance under these settings, models optimized for in-domain accuracy may rely on superficial lexical or dataset-specific cues, resulting in reduced robustness under distributional shift.
In this study, we introduce an evaluation framework based on controlled semantic perturbations to assess the robustness of a publication type classifier and investigate robustness-oriented training strategies that combine entity masking and domain-adversarial training to mitigate reliance on spurious topical correlations.
Our results show that the commonly observed trade-off between robustness and in-domain accuracy can be mitigated when robustness objectives are designed to selectively suppress non–task-defining features while preserving salient methodological signals. We find that these improvements arise from two complementary mechanisms: (1) increased reliance on explicit methodological cues when such cues are present in the input, and (2) reduced reliance on spurious domain-specific topical features. These findings highlight the importance of feature-level robustness analysis for publication type and study design classification and suggest that refining masking and adversarial objectives to more selectively suppress topical information may further improve robustness. Data, code, and models are available at \url{https://github.com/ScienceNLP-Lab/MultiTagger-v2/tree/main/ICHI}.
\end{abstract}

\begin{IEEEkeywords}
\textit{biomedical publication type classification, robustness evaluation, domain-adversarial training, entity masking}
\end{IEEEkeywords}

\section{Introduction}
With the rapid growth of biomedical literature added to PubMed every day, efficiently accessing relevant information has become increasingly challenging for researchers and clinicians seeking to stay current with emerging biomedical knowledge. Biomedical publications contain vast amounts of information that underpin key downstream applications such as knowledge discovery and evidence synthesis. Accurate and consistent indexing of these publications is therefore a fundamental prerequisite for enabling such research and facilitating informed clinical decisions. The U.S. National Library of Medicine (NLM) has initiated efforts to automatically index articles in MEDLINE using both the Medical Subject Headings (MeSH) thesaurus and publication type (PT) metadata \cite{indexing_nlm}. Over the decades, substantial progress has been made through NLM’s Medical Text Indexer \cite{aronson2004nlm} program toward more consistent and accurate MeSH indexing. Early approaches relied on term matching, followed by semantic search methods that aimed to disambiguate synonyms and abbreviations of complex biomedical terms \cite{krithara2023road}. More recently, fine-tuning transformer-based models such as PubMedBERT \cite{gu2021domain} has become the dominant paradigm, exemplified by systems like BERTMeSH \cite{you2021bertmesh}, which leverage rich contextualized representations learned from large-scale biomedical corpora to achieve superior indexing performance.

While MeSH topical indexing has received extensive attention and is well suited for retrieving biomedical literature by subject matter, indexing biomedical publications by methodological design, which reflects how a study was conducted (e.g., Randomized Controlled Trials, Cohort Studies), as well as by publication type (e.g., Case Reports), remains comparatively underexplored despite its importance. Automatically tagging publication types and study designs (collectively referred to as PTs here) can accelerate method-oriented literature retrieval, as well as downstream evidence synthesis tasks such as systematic reviews and meta-analyses, where reviewers often spend countless hours manually screening articles \cite{cohen2010evidence, schneider2022evaluation}. For example, the RCT Tagger \cite{cohen2015automated} helps identify whether an article describes a randomized controlled trial. Such automatic identification is valuable for retrieving primary evidence in Evidence-Based Medicine (EBM) \cite{sackett1996evidence}. Building on this line of work, prior research introduced MultiTagger, a SVM-based multi-label system that assigns predictive scores to 50 PTs \cite{cohen2021fifty}. More recent studies further extend this direction using transformer-based encoder models \cite{menke2025publication, menke2025enhancing}.

Prior studies have shown that deep learning models often rely on spurious correlations or surface-level cues present in the training data, limiting their ability to generalize to unseen distributions where such associations no longer hold \cite{murali2023beyond}. Model robustness in publication type and study design classification is therefore essential for real-world deployment, where the same underlying study design may appear across clinical domains (e.g., podiatry vs.\ ophthalmology) and be described using different biomedical terminology over time.

Robustness evaluation encompasses multiple notions depending on the application context and the types of distributional variation considered \cite{sun2025role}. In PT classification, labels are intended to reflect the methodological properties of a study, rather than its topical biomedical content or specific entity mentions. Following prior work that conceptualizes robustness as invariance of model predictions under text perturbations \cite{wang2022measure}, we define robustness in this study as \textit{a PT classifier's ability to preserve correct predictions across topical shifts by relying on method-specific signals rather than topical shortcuts.}

\begin{figure}[]
\centering
\includegraphics[width=\columnwidth]{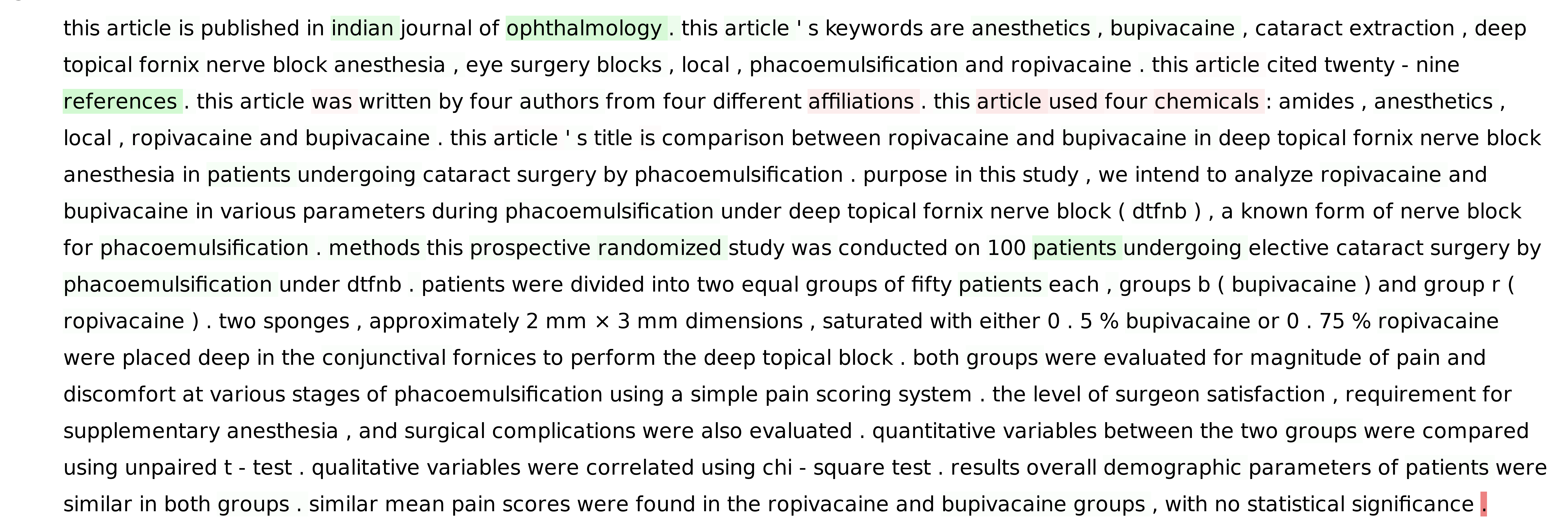}
\caption{Saliency visualization of the baseline model for PMID 30127137, illustrating a false-negative prediction for the \textit{Randomized Controlled Trial Humans} (predicted probability = 0.452).}
\label{intro}
\end{figure}

To assess the extent to which the model reported in Menke et al. \cite{menke2025enhancing} leverages methodological signals in its predictions, we analyze token-level importance attributions using gradient-based saliency mapping \cite{bastings2020elephant}, implemented via Captum \cite{kokhlikyan2020captum}. Figure \ref{intro} presents an example in which a large proportion of attention allocated to topical terms dilutes the relative contribution of core methodological indicators (e.g., ``randomized'' and ``prospective''), leading to false-negative predictions even when randomized study designs are explicitly stated.
This behavior motivates a systematic investigation of the robustness of the existing PT classifier and the development of training strategies that encourage models to prioritize method-defining signals over topical correlations.

Our study investigates the following research questions:
\begin{enumerate}
    \item RQ1: How sensitive are PT classifiers to controlled semantic shifts that preserve study design and publication type semantics?
    \item RQ2: Can we reduce reliance on domain-specific lexical cues to improve robustness without degrading in-domain predictive performance?
    \item RQ3: Which robustness-oriented training strategy, domain-adversarial training, masked-entity training, or their combination, achieves the best trade-off between robustness and in-domain performance?
\end{enumerate}

Our contributions are:
\begin{enumerate}
    \item To address RQ1, we propose a knowledge-guided semantic perturbation framework to systematically evaluate the sensitivity of PT classifiers under controlled semantic shifts that preserve methodological meaning.
    \item To address RQ2, we develop and evaluate robustness-oriented training strategies and examine whether reducing reliance on domain-specific lexical cues improves robustness to semantic perturbations without degrading predictive performance on clean, in-domain data, relative to the model reported in Menke et al. \cite{menke2025enhancing} (referred to as \textsc{baseline}).
    \item To address RQ3, we analyze how domain-adversarial training and masked-entity training influence model reliance on topical versus methodological features, and evaluate whether their combination can mitigate the robustness–accuracy trade-off in PT classification.

\end{enumerate}

\section{Related Work}

Evaluating the robustness of biomedical NLP models has gained increasing attention in recent years. Early studies examined the robustness of BioBERT, PubMedBERT, and other domain-specific models across multiple biomedical NLP tasks under adversarial and noisy perturbations, demonstrating that even small word- or character-level changes can lead to substantial performance degradation \cite{moradi2022improving}. Similarly, prior work showed that substituting medical synonyms in biomedical named entity recognition and semantic textual similarity tasks results in significant accuracy drops \cite{araujo2020adversarial}.

Adversarial training has been widely explored to improve the robustness of NLP models against distributional shifts and input perturbations \cite{goyal2023survey}. For example, one study \cite{moradi2022improving} augmented training data with a mix of noisy and entity-swapped adversarial examples alongside clean inputs. Models trained on this augmented data demonstrated improved robustness to perturbed test inputs and achieved a 2\% improvement in accuracy on clean data. Following a similar idea, entity-masking strategies have been explored as complementary approaches to reduce model reliance on spurious lexical cues. For instance, Pergola et al. \cite{pergolaboosting} fine-tuned biomedical QA models using an entity-aware masking strategy, where entity mentions were replaced with their semantic types or generic placeholders. 
Another line of work introduces a secondary domain classifier trained jointly with the main task classifier via a gradient reversal layer, encouraging learned representations to become indistinguishable across domains and thus enforcing domain invariance \cite{ganin2015unsupervised}.

Prior work on robustness in NLP has primarily focused on mitigating spurious correlations arising from domain shifts, such as changes in data source, genre, or time \cite{calderon2024measuring}, or from surface-level lexical variation introduced through synonym substitution or noise-based perturbations \cite{wang2021robustness}. While these studies evaluate whether model predictions remain stable under distributional or lexical changes, they generally do not explicitly analyze which input features or signals the model relies on to produce and maintain correct predictions.

In contrast, our work focuses on a challenge specific to PT classification: method-defining signals are often entangled with topic-specific entities that may be statistically correlated with target labels but are not necessarily indicative of study design. In addition to evaluating prediction stability, we analyze how robustness-oriented training redistributes model attention between method-defining cues and topic-specific features, providing a feature-level explanation for when and why robustness improvements occur.

\begin{table*}[h]
\caption{Three types of perturbations applied to the original text (PMID: 10050264).}
\centering
\scriptsize
\begin{tabularx}{\textwidth}{l X}
\toprule
\textbf{Operation} & \textbf{Sentence} \\
\midrule
Original &
Dietary factors are widely studied as risk factors for colorectal cancer,
with much information from case-control studies. \\

Synonym substitution \\
(e.g., colorectal cancer $\rightarrow$ large intestine cancer) &
Dietary factors are widely studied as risk factors for large intestine cancer,
with much information from case-control studies. \\

Concept substitution \\
(e.g., colorectal cancer $\rightarrow$ an advanced bladder cancer) &
Dietary factors are widely studied as risk factors for
\textit{Malignant neoplasm of dome of urinary bladder stage IV},
with much information from case-control studies. \\

Synonym / Concept substitution + EDA \\
(``Dietary'' and ``bladder stage'' were deleted) &
Factors are widely studied as risk factors for
\textit{Malignant neoplasm of dome urinary IV},
with much information from case-control studies. \\
\bottomrule
\end{tabularx}
\label{perturbation}
\end{table*}

\section{Materials and Methods}

\subsection{Task formulation}
Given an input document $x$ (e.g., a biomedical article abstract), the model predicts a set of PT labels $y$. Models are trained and evaluated on samples $(x, y) \sim \mathcal{D}$, and robustness is assessed by evaluating model behavior on perturbed inputs $(x', y) \sim \mathcal{D}' \neq \mathcal{D}$ that preserve task-relevant semantics, following standard robustness evaluation settings in \cite{wang2022measure, goyal2023survey}. 

\subsection{Dataset}
The dataset, originally introduced by Menke et al. \cite{menke2025enhancing}, contains 166,192 articles, split into training (70\%, $n=116{,}368$), validation (10\%, $n=16{,}619$), and test (20\%, $n=33{,}205$) sets, with similar class distributions across all splits. Each instance is constructed by verbalizing multiple document-level features extracted from PubMed (e.g., title, journal, keywords, and chemicals) and concatenating them with the article title and abstract.

\subsection{Baseline Model}
The baseline model is built on SPECTER2-base \cite{singh-etal-2023-scirepeval} and optimized for multi-label classification using asymmetric loss (ASL) \cite{ridnik2021asymmetric} with label smoothing \cite{muller2019does} to address class imbalance and improve model calibration. To enhance representation learning, the training objective further incorporates HeroCon \cite{zheng2022contrastive}, a supervised contrastive learning loss, and an unsupervised contrastive loss, ADjusted InfoNCE (ADNCE) \cite{wu2024understanding}, together with the primary label loss (ASL) via a weighted aggregation. This model serves as the baseline for our analysis and subsequent improvements. All baseline hyperparameters follow the configuration reported in Menke et al. \cite{menke2025enhancing}. In the following sections, we first evaluate the baseline model's performance under perturbation and then introduce three training strategies to mitigate spurious correlations and improve robustness.

\begin{figure*}[ht]
    \centering
    \includegraphics[width=\textwidth]{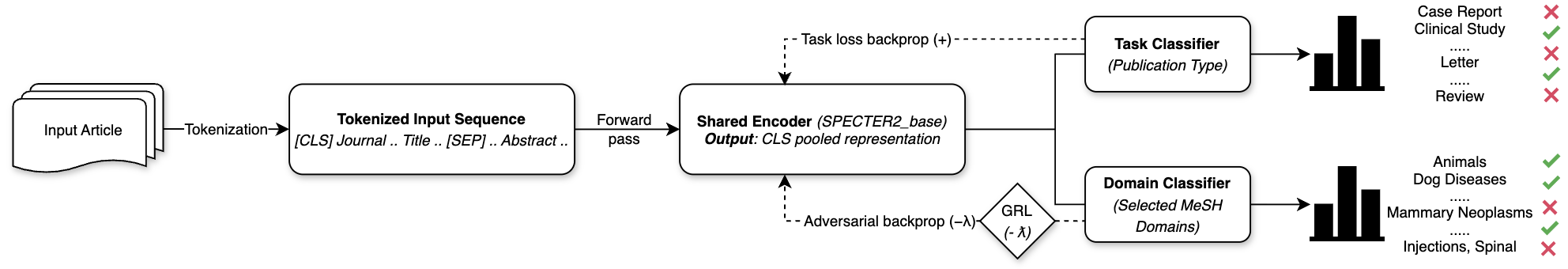}
    \caption{Overview of the adversarial training architecture. A pre-trained encoder (SPECTER2-base) produces shared representations that are fed into a PT classifier and an auxiliary domain classifier. The PT classifier predicts probabilities over 61 labels. The domain classifier predicts 53 selected MeSH terms. Gradients from the domain classifier are reversed through a Gradient Reversal Layer (GRL) to suppress domain-specific signals.}
    \label{adversarial_training}
\end{figure*}

\subsection{Knowledge-Guided Perturbations for Robustness Evaluation}

To evaluate model robustness, we construct perturbed test sets that systematically modify the original inputs while preserving underlying study-design semantics and PT labels. Rather than applying generic noise-based perturbations (e.g., typos or arbitrary word swaps), we use UMLS-guided substitutions to modify topical biomedical content while maintaining PT-defining methodological meaning.

Specifically, we restrict perturbations to a subset of UMLS semantic groups \cite{bodenreider2004unified} that primarily capture topical biomedical content and are unlikely to affect PT labels, namely \textit{Chemicals and Drugs, Anatomy, Disorders, Genes and Molecular Sequences, Geographic Areas, and Physiology}. Entities belonging to these groups are identified using MetaMap \cite{aronson2010overview} and replaced with synonym sets or alternative biomedical concepts from the UMLS Metathesaurus \cite{bodenreider2004unified}.

We consider three primary perturbation operations, ordered from mild to increasingly severe. Examples of each perturbation are shown in Table \ref{perturbation}. 
\begin{enumerate}
    \item Synonym substitution: replaces an entity with a synonym referring to the same biomedical concept.
    \item Concept substitution: replaces an entity with a different biomedical concept of the same UMLS semantic type, modifying topical content while maintaining the entity's semantic role in the sentence.
    \item Synonym / Concept substitution + EDA: further introduces lexical and syntactic noise using simple Easy Data Augmentation (EDA) operations \cite{wei2019eda}, such as random word deletion, insertion, or swapping, applied on top of synonym/concept substitution. In extreme cases, EDA may degrade sentence readability and fluency beyond topical variation; however, they are constructed such that the underlying PT semantics, and therefore the ground-truth label, remain unchanged.
\end{enumerate}

Perturbations were applied to the entire constructed input at varying magnitudes, replacing 30\%, 50\%, or 100\% of the identified entities. To evaluate robustness under the most challenging setting, we adopt concept substitution applied to all entities (100\%), combined with Easy Data Augmentation (EDA), as the primary perturbed test set.

Overall, we design controlled perturbations that introduce lexical, semantic, and syntactic variation while preserving PT-defining semantics. A robust model should therefore maintain stable predictions despite changes in surface form and topical expression.

\subsection{Models for Robust PT Classification}
Below, we describe three knowledge-guided training strategies aimed at improving robustness beyond the baseline model: (1) masked-entity training, (2) domain-adversarial training, and (3) a joint training strategy that integrates entity masking with domain-adversarial training.

\subsubsection{Masked-Entity Training}
For masked-entity training, denoted as \textsc{Mask} in the following sections, we preprocess the input by identifying biomedical entities using MetaMap and selecting entities belonging to UMLS semantic groups that are unlikely to affect study design, namely \textit{Chemicals and Drugs, Anatomy, Disorders, Genes and Molecular Sequences, Geographic Areas, and Physiology}. These entities were replaced with typed markers (e.g., \textsc{[chemical]}, \textsc{[anatomy]}, \textsc{[disorder]}, \textsc{[gene]}, \textsc{[geographic]}, \textsc{[physiology]}), and the tokenizer vocabulary was updated accordingly to jointly fine-tune the embeddings of these markers with the model. This masked-entity representation standardizes specific biomedical entity mentions and encourages the model to focus on PT–related patterns rather than domain-specific content such as diseases or drugs.

Prior work has shown that augmenting training data with a mixture of noisy and entity-swapped adversarial examples can improve robustness \cite{moradi2022improving}. Analogous to this setup, we construct augmented training data by randomly selecting 50\% of training instances for entity masking while retaining the original text for the remaining instances, thereby exposing the model to both masked and unmasked inputs during training. All other model architectures and configurations, including loss functions and hyperparameters, remain identical to the baseline model \cite{menke2025enhancing}.

\subsubsection{Domain-Adversarial Training}
For domain-adversarial training, denoted as \textsc{Adversarial} in the following sections, we adopt the Domain-Adversarial Neural Network (DANN) framework \cite{ganin2016domain}, in which a shared encoder is optimized for the primary task while learning representations that are invariant to domain-specific information via adversarial gradient reversal from an auxiliary domain classifier. 

As illustrated in Figure \ref{adversarial_training}, the model consists of a shared encoder, a task classifier, and an auxiliary domain classifier. Given an input document, the encoder produces a pooled \textsc{[CLS]} representation, which is used as input to both the task classifier and the domain classifier. A gradient reversal layer (GRL) is placed between the encoder and the domain classifier. During backpropagation, the GRL scales the gradients from the domain classifier by a factor of $-\lambda(p)$, where $\lambda(p)$ controls the strength of the domain-adversarial signal as a function of training progress $p$. Specifically, $\lambda(p)$ is defined as:

{\scriptsize
\[
\lambda(p) = \lambda_{\max} \left( \frac{2}{1 + e^{-10p}} - 1 \right),
\quad
p = \frac{\text{epoch}}{\text{max\_epoch}}
\]
}

The domain-adversarial training objective minimizes the following total loss, where the task loss follows the baseline configuration of Menke et al. \cite{menke2025enhancing}:

{\scriptsize
\begin{equation}
\mathcal{L}_{\text{total}}
=
\underbrace{
\mathcal{L}_{\text{ASL}}
+
\alpha \mathcal{L}_{\text{unsupCL}}
+
\beta \mathcal{L}_{\text{supCL}}
}_{\mathcal{L}_{\text{task}}}
\;+\;
\underbrace{
\lambda(p)\mathcal{L}_{\text{domain}}
}_{\mathcal{L}_{\text{domain}}}
\end{equation}
}

Here, $\mathcal{L}_{\text{ASL}}$ denotes the asymmetric loss with label smoothing for multi-label classification, while $\mathcal{L}_{\text{unsupCL}}$ and $\mathcal{L}_{\text{supCL}}$ denote the unsupervised and supervised contrastive losses, weighted by hyperparameters $\alpha$ and $\beta$, respectively. The domain loss $\mathcal{L}_{\text{domain}}$ corresponds to the auxiliary domain classification objective, scaled by the scheduling function $\lambda(p)$.

To construct domain labels, we use MeSH terms associated with each article. Since certain MeSH terms encode publication type information or methodological characteristics that are directly relevant to the primary task objective, we exclude MeSH terms belonging to the following branches when constructing the auxiliary domain classifier. Specifically, we excluded MeSH branches that contain PT labels in the MeSH hierarchy, namely \textit{E: Analytical, Diagnostic and Therapeutic Techniques}, \textit{H: Disciplines and Occupations}, \textit{I: Anthropology, Education, Sociology, and Social Phenomena}, \textit{L: Information Science}, \textit{N: Health Care}, and \textit{V: Publication Characteristics}, to avoid leaking task-relevant information into the domain classifier. For example, category E contains MeSH descriptors such as \textit{Double-Blind Method} and \textit{Follow-Up Studies}, which are used as labels in our PT classification task.

From the remaining MeSH terms, we compute term frequencies over the training corpus and retain terms appearing in at least 1\% of the documents, resulting in 53 domain labels (e.g., \textit{Animals, Cardiovascular Diseases}). This threshold ensures sufficient coverage of domain-related features while avoiding label sparsity, thereby providing more stable gradient signals for adversarial training.

\subsubsection{Adversarial Training with Masked Input}
In the joint training setup, half of the training instances are randomly selected for entity masking using the same masking ratio as in the masked-entity setting, while domain-adversarial training is applied to all training instances as an auxiliary objective. This combination is denoted as \textsc{Mask + Adversarial}.

\subsection{Experimental Setup}
We developed and compared different fine-tuning strategies and evaluated all fine-tuned models on both the original (clean) and perturbed test sets. All hyperparameters were tuned on the validation set, and the checkpoint with the highest performance on validation was saved for final evaluation.

We adopted the same hyperparameter configuration as reported in Menke et al. \cite{menke2025enhancing}, including a batch size of 32, a learning rate of $1\times10^{-4}$ for transformer layers and $1\times10^{-2}$ for the classification layer, the RAdam optimizer, a dropout rate of 0.1, and label smoothing with $\alpha = 0.05$. The asymmetric loss was parameterized with $\gamma_{-}=4$, $\gamma_{+}=1$, and $m=0.05$, and the supervised contrastive loss weight was set to $\beta = 0.1$. Early stopping was applied if no improvement in validation macro-F1 was observed for 4 consecutive epochs, with a maximum of 25 training epochs. For domain-adversarial training, we set the maximum adversarial weight to $\lambda_{\max} = 0.5$. All experiments were conducted on a single Tesla V100 GPU with 32GB of memory. Each training run for the baseline and masking models takes approximately 11.25 hours, while training with the adversarial layer requires approximately 19.8 hours. All models were implemented using PyTorch (v2.5.1).

\subsection{Evaluation Metrics}
We evaluate model performance on both clean and perturbed test sets to compare the baseline model with the robustness-oriented training strategies along two dimensions: \emph{task performance} and \emph{robustness}. \emph{Task performance} is assessed on the original clean test set using standard multi-label classification metrics, reflecting how well each model predicts PTs under in-domain conditions. These metrics include precision, recall, and $F_1$ scores, as well as the area under the precision-recall curve (AUPRC), with both micro and macro averaging. To assess probabilistic calibration, we additionally measure Expected Calibration Error (ECE) under $\ell_1$, $\ell_2$, and $\ell_{\max}$ norms \cite{naeini2015obtaining}, where lower values indicate better calibration. We assess statistical significance using paired one-sided $t$-tests over 1,000 bootstrap resamples of the test set.

We evaluate robustness using complementary metrics commonly adopted in text classification and adaptable to our setting:

\begin{enumerate}
    \item \emph{Robust accuracy} \cite{morris2020textattack, goyal2023survey}, measured as the absolute $F_1$ score on perturbed data;
    \item \emph{Performance degradation} \cite{kumar2025robustness, goyal2023survey}, defined as the difference between performance on clean and perturbed test sets where smaller performance drops indicate stronger robustness;
    \item \emph{Attack success rate (ASR)} \cite{goyal2023survey, kumar2025robustness}, quantifies the proportion of predictions that are correct on clean inputs but become incorrect (i.e., false negatives) under perturbation, aggregated across labels and instances. This metric is conditioned on correctness on the clean input; higher ASR indicates lower robustness. ASR is defined as:
    
    {\scriptsize
    \[
\mathrm{ASR}_{\text{micro}}=
\frac{\sum_i \left| (Y_i \cap \hat{Y}_i^{\text{clean}})\setminus \hat{Y}_i^{\text{perturbed}} \right|}
{\sum_i \left| Y_i \cap \hat{Y}_i^{\text{clean}}\right|}\times 100
\]
}

where $Y_i$ denotes the gold label set for instance $i$, 
$\hat{Y}_i^{\text{clean}}$ and $\hat{Y}_i^{\text{perturbed}}$ denote the predicted label sets on clean and perturbed inputs, respectively. 

\end{enumerate}

In addition, we report macro-averaged performance over the 22 core PTs considered most critical for evidence synthesis (e.g., \textit{Cohort Studies}, \textit{Case–Control Studies}), as defined in Menke et al. \cite{menke2025enhancing}.

\section{Results}

\newcommand{\twoline}[2]{\makecell[c]{\small #1 \\ \scriptsize #2}}

\begin{table*}[t]
\caption{Robust accuracy under extreme semantic perturbation (100\% concept substitution + EDA). Micro and Macro denote averaging over all PTs; Core denotes macro-averaged performance over the 22 core PTs. 95\% CIs are computed via bootstrap sampling ($n=1{,}000$, test set $n=33{,}205$).}
\centering

\setlength{\tabcolsep}{4pt}
\renewcommand{\arraystretch}{1.25}

\resizebox{\textwidth}{!}{
\begin{tabular}{lccccccccccccccc}
\toprule

& \multicolumn{3}{c}{\textbf{$F_1$}}
& \multicolumn{3}{c}{\textbf{Precision}}
& \multicolumn{3}{c}{\textbf{Recall}}
& \multicolumn{3}{c}{\textbf{AUPRC}}
& \multicolumn{3}{c}{\textbf{ECE $\downarrow$}} \\

\cmidrule(lr){2-4}
\cmidrule(lr){5-7}
\cmidrule(lr){8-10}
\cmidrule(lr){11-13}
\cmidrule(lr){14-16}

\textbf{Model}
& Micro & Macro & Core
& Micro & Macro & Core
& Micro & Macro & Core
& Micro & Macro & Core
& $\ell_1$ & $\ell_2$ & $\ell_{\max}$ \\

\midrule

\textsc{baseline}
& \twoline{0.573}{[0.569--0.576]}
& \twoline{0.573}{[0.564--0.581]}
& \twoline{0.570}{[0.565--0.575]}
& \textbf{\twoline{0.649}{[0.644--0.652]}}
& \textbf{\twoline{0.691}{[0.675--0.702]}}
& \textbf{\twoline{0.679}{[0.673--0.684]}}
& \twoline{0.513}{[0.510--0.517]}
& \twoline{0.525}{[0.517--0.533]}
& \twoline{0.525}{[0.520--0.531]}
& \twoline{0.637}{[0.633--0.640]}
& \twoline{0.620}{[0.611--0.629]}
& \twoline{0.632}{[0.626--0.638]}
& \twoline{0.166}{[0.165--0.166]}
& \twoline{0.182}{[0.182--0.182]}
& \twoline{0.292}{[0.290--0.293]} \\

\addlinespace[4pt]

\textsc{mask}
& \twoline{0.601}{[0.598--0.605]}
& \twoline{0.604}{[0.595--0.612]}
& \twoline{0.601}{[0.597--0.606]}
& \twoline{0.647}{[0.644--0.651]}
& \twoline{0.685}{[0.671--0.696]}
& \twoline{0.666}{[0.661--0.671]}
& \twoline{0.562}{[0.558--0.565]}
& \twoline{0.567}{[0.559--0.576]}
& \twoline{0.576}{[0.571--0.581]}
& \twoline{0.659}{[0.655--0.662]}
& \twoline{0.644}{[0.636--0.653]}
& \twoline{0.656}{[0.651--0.662]}
& \twoline{0.183}{[0.182--0.183]}
& \twoline{0.199}{[0.199--0.199]}
& \twoline{0.315}{[0.313--0.317]} \\

\addlinespace[4pt]

\textsc{adversarial}
& \twoline{0.576}{[0.573--0.579]}
& \twoline{0.584}{[0.575--0.593]}
& \twoline{0.570}{[0.565--0.575]}
& \twoline{0.627}{[0.623--0.631]}
& \twoline{0.682}{[0.666--0.693]}
& \twoline{0.673}{[0.668--0.679]}
& \twoline{0.533}{[0.529--0.536]}
& \twoline{0.542}{[0.534--0.550]}
& \twoline{0.536}{[0.530--0.541]}
& \twoline{0.646}{[0.642--0.649]}
& \twoline{0.629}{[0.619--0.638]}
& \twoline{0.637}{[0.631--0.642]}
& \textbf{\twoline{0.162}{[0.162--0.162]}}
& \textbf{\twoline{0.179}{[0.179--0.179]}}
& \textbf{\twoline{0.291}{[0.290--0.293]}} \\

\addlinespace[4pt]

\textsc{mask + adversarial}
& \textbf{\twoline{0.604}{[0.601--0.607]}}
& \textbf{\twoline{0.606}{[0.597--0.614]}}
& \textbf{\twoline{0.604}{[0.599--0.609]}}
& \twoline{0.637}{[0.634--0.641]}
& \twoline{0.678}{[0.665--0.690]}
& \twoline{0.668}{[0.663--0.674]}
& \textbf{\twoline{0.575}{[0.571--0.578]}}
& \textbf{\twoline{0.572}{[0.563--0.580]}}
& \textbf{\twoline{0.578}{[0.572--0.583]}}
& \textbf{\twoline{0.664}{[0.660--0.668]}}
& \textbf{\twoline{0.645}{[0.637--0.654]}}
& \textbf{\twoline{0.658}{[0.653--0.664]}}
& \twoline{0.176}{[0.175--0.176]}
& \twoline{0.193}{[0.192--0.193]}
& \twoline{0.308}{[0.307--0.310]} \\

\bottomrule
\end{tabular}
}
\label{delta_extreme}
\end{table*}

\begin{table}[t]
\caption{Attack Success Rate (ASR) and absolute micro-averaged performance degradation ($\Delta$) relative to clean performance under extreme semantic perturbation.}
\centering
\scriptsize
\begin{adjustbox}{width=\columnwidth}
\begin{tabular}{l|c|c|c|c|c}
\hline
\textbf{Model} & \textbf{ASR (\%)} & \textbf{$\Delta F_1$} & \textbf{$\Delta$Precision} & \textbf{$\Delta$Recall}  & \textbf{$\Delta$AUPRC} \\
\hline
\textsc{Baseline} & 25.41 & -0.095 & -0.028 & -0.145 & -0.099 \\
\textsc{Mask} & 20.55 & \textbf{-0.067} & \textbf{-0.009} & -0.118 & \textbf{-0.077} \\
\textsc{Adversarial} & 25.31 & -0.095 & -0.039 & -0.144 & -0.095 \\
\textsc{Mask + Adversarial} & \textbf{19.94} & \textbf{-0.067} & -0.015 & \textbf{-0.116} & \textbf{-0.077} \\
\hline
\end{tabular}
\label{asr}
\end{adjustbox}
\end{table}

\subsection{Baseline Model Robustness}
We first evaluate the robustness of the baseline PT classifier under controlled semantic perturbations. Table~\ref{delta_extreme} reports \textit{robust accuracy}, and Table~\ref{asr} summarizes the \textit{attack success rate (ASR)} and \textit{performance degradation} relative to clean test performance. The baseline model exhibits substantial performance degradation under extreme perturbation, with a drop of 0.095 in micro-averaged $F_1$ and 0.145 in recall. An ASR of 25.41\% indicates that a substantial fraction of predictions that are correct on clean inputs become incorrect after perturbation.

\subsection{Improving Robustness with Robustness-Oriented Training}

Next, we evaluate robustness-oriented training strategies under extreme semantic perturbation. Results are reported in Table \ref{delta_extreme}. For masked-entity training, the default masking ratio is 50\% of training instances unless otherwise specified. The \textsc{Mask + Adversarial} model achieves the highest \textit{robust accuracy}, with a micro-$F_1$ of 0.604 and a macro-$F_1$ of 0.606, followed by the \textsc{Mask} model (micro-$F_1$ = 0.601; macro-$F_1$ = 0.604). The \textsc{Mask + Adversarial} model also achieves higher recall under perturbation (micro = 0.575; macro = 0.572) and the highest AUPRC, with micro- and macro-averaged scores of 0.664 and 0.645, respectively. The \textsc{Adversarial} model yields the lowest expected calibration error across all metrics ($\ell_1$ = 0.162, $\ell_2$ = 0.179, $\ell_{\max}$ = 0.291).
In terms of \textit{performance degradation}, as shown in Table \ref{asr}, both the \textsc{Mask + Adversarial} and \textsc{Mask} models exhibit the smallest drop in micro-$F_1$ ($\Delta = -0.067$). Among all models, the \textsc{Mask + Adversarial} model achieves the lowest attack success rate, followed by the \textsc{Mask} model.

\begin{figure}[h]
    \centering
    \includegraphics[width=0.48\textwidth]{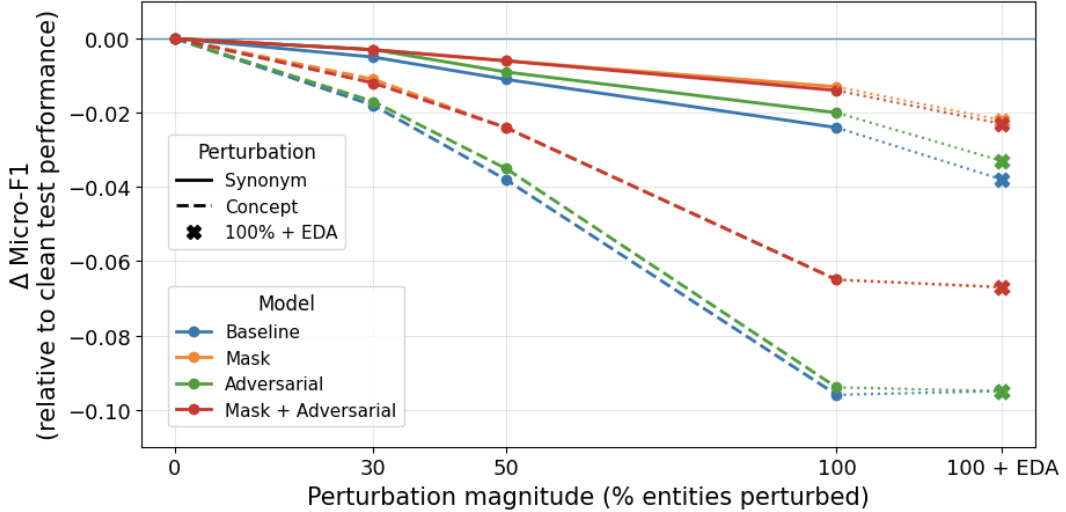}
    \caption{Changes in micro-$F_1$ ($\Delta$) under increasing levels of semantic perturbation, measured relative to clean test performance. Solid and dashed lines indicate synonym and concept substitutions, respectively.}
    \label{degradation}
\end{figure}

\subsection{Performance Degradation under Increasing Perturbation Magnitude}

Figure \ref{degradation} summarizes changes in micro-$F_1$ ($\Delta$) under increasing perturbation severity, including synonym and concept substitution applied at replacement ratios of 30\%, 50\%, and 100\%, as well as additional syntactic noise introduced via EDA. Across all models, performance degrades as perturbation magnitude increases, with larger drops observed under concept substitution than synonym substitution.

The baseline model exhibits the steepest performance degradation under both synonym and concept substitution, particularly at 100\% replacement. In contrast, robustness-oriented training strategies degrade more gradually and less than the baseline. At 30\% synonym substitution, performance degradation curves partially overlap across models, indicating limited sensitivity to mild lexical variation. At higher perturbation levels (50–100\%), the \textsc{Mask} and \textsc{Mask + Adversarial} models show consistently smaller declines in micro-$F_1$ than the baseline and the \textsc{Adversarial} model, with the strongest robustness observed under 100\% concept substitution.

\subsection{Performance on the Clean Test Set Across Robustness-Oriented Training Strategies}
We evaluate model performance on the clean test set to assess whether robustness-oriented training strategies preserve or improve in-domain accuracy. As shown in Table~\ref{cleandata_comparsion}, both the \textsc{Adversarial} and \textsc{Mask + Adversarial} models achieve micro- and macro-averaged $F_1$ scores slightly higher than the baseline. The \textsc{Adversarial} model achieves the highest performance (micro-$F_1$ = 0.671; macro-$F_1$ = 0.677), while the \textsc{Mask + Adversarial} model achieves the same micro-$F_1$ and a slightly lower macro-$F_1$ (0.676).

\begin{table*}[th]
\caption{Model performance on the original (clean) test set. Micro and Macro denote averaging over all PTs; Core denotes macro-averaged performance over the 22 core PTs. 95\% CIs are computed via bootstrap sampling ($n=1{,}000$, test set $n=33{,}205$).}
\centering

\setlength{\tabcolsep}{4pt}
\renewcommand{\arraystretch}{1.25}

\resizebox{\textwidth}{!}{
\begin{tabular}{lccccccccccccccc}
\toprule

& \multicolumn{3}{c}{\textbf{$F_1$}}
& \multicolumn{3}{c}{\textbf{Precision}} 
& \multicolumn{3}{c}{\textbf{Recall}}
& \multicolumn{3}{c}{\textbf{AUPRC}}
& \multicolumn{3}{c}{\textbf{ECE $\downarrow$}} \\

\cmidrule(lr){2-4}
\cmidrule(lr){5-7}
\cmidrule(lr){8-10}
\cmidrule(lr){11-13}
\cmidrule(lr){14-16}

\textbf{Model}
& Micro & Macro & Core 
& Micro & Macro & Core 
& Micro & Macro & Core 
& Micro & Macro & Core
& $\ell_1$ & $\ell_2$ & $\ell_{\max}$ \\

\midrule

\textsc{baseline} 
& \twoline{0.668}{[0.665--0.670]} 
& \twoline{0.668}{[0.661--0.675]} 
& \textbf{\twoline{0.689}{[0.685--0.694]}} 
& \textbf{\twoline{0.677}{[0.674--0.681]}} 
& \textbf{\twoline{0.701}{[0.692--0.710]}} 
& \textbf{\twoline{0.711}{[0.706--0.716]}} 
& \twoline{0.658}{[0.655--0.662]} 
& \twoline{0.658}{[0.650--0.665]} 
& \twoline{0.676}{[0.670--0.681]} 
& \twoline{0.736}{[0.732--0.739]} 
& \twoline{0.712}{[0.705--0.720]} 
& \twoline{0.735}{[0.730--0.740]} 
& \twoline{0.157}{[0.156--0.157]} 
& \twoline{0.179}{[0.179--0.180]} 
& \twoline{0.320}{[0.318--0.322]} \\

\addlinespace[4pt]

\textsc{mask} 
& \twoline{0.668}{[0.665--0.671]} 
& \twoline{0.675}{[0.667--0.681]} 
& \twoline{0.685}{[0.681--0.689]} 
& \twoline{0.656}{[0.653--0.659]} 
& \twoline{0.689}{[0.680--0.698]} 
& \twoline{0.687}{[0.683--0.693]} 
& \twoline{0.680}{[0.677--0.683]} 
& \twoline{0.681}{[0.674--0.686]} 
& \twoline{0.694}{[0.689--0.699]} 
& \twoline{0.736}{[0.733--0.740]} 
& \twoline{0.718}{[0.711--0.724]} 
& \twoline{0.733}{[0.728--0.738]} 
& \twoline{0.170}{[0.170--0.170]} 
& \twoline{0.192}{[0.192--0.192]} 
& \twoline{0.335}{[0.333--0.337]} \\

\addlinespace[4pt]

\textsc{adversarial} 
& \textbf{\twoline{0.671}{[0.669--0.674]}} 
& \textbf{\twoline{0.677}{[0.670--0.684]}} 
& \textbf{\twoline{0.689}{[0.684--0.693]}} 
& \twoline{0.666}{[0.663--0.669]} 
& \twoline{0.690}{[0.681--0.700]} 
& \twoline{0.699}{[0.694--0.704]} 
& \twoline{0.677}{[0.674--0.680]} 
& \twoline{0.678}{[0.671--0.685]} 
& \twoline{0.687}{[0.682--0.692]} 
& \textbf{\twoline{0.741}{[0.737--0.744]}} 
& \twoline{0.720}{[0.713--0.727]} 
& \textbf{\twoline{0.736}{[0.731--0.741]}} 
& \textbf{\twoline{0.154}{[0.154--0.154]}} 
& \textbf{\twoline{0.176}{[0.176--0.177]}} 
& \textbf{\twoline{0.318}{[0.315--0.320]}} \\

\addlinespace[4pt]

\textsc{mask + adversarial} 
& \textbf{\twoline{0.671}{[0.668--0.673]}} 
& \twoline{0.676}{[0.668--0.683]} 
& \twoline{0.687}{[0.683--0.692]} 
& \twoline{0.652}{[0.648--0.655]} 
& \twoline{0.684}{[0.674--0.694]} 
& \twoline{0.688}{[0.683--0.693]} 
& \textbf{\twoline{0.691}{[0.687--0.694]}} 
& \textbf{\twoline{0.684}{[0.677--0.691]}} 
& \textbf{\twoline{0.697}{[0.692--0.702]}} 
& \textbf{\twoline{0.741}{[0.737--0.744]}} 
& \textbf{\twoline{0.721}{[0.714--0.727]}} 
& \textbf{\twoline{0.736}{[0.731--0.741]}} 
& \twoline{0.164}{[0.164--0.165]} 
& \twoline{0.186}{[0.185--0.186]} 
& \twoline{0.330}{[0.328--0.332]} \\

\bottomrule
\end{tabular}
}
\label{cleandata_comparsion}
\end{table*}

Paired one-sided bootstrap significance testing with 1,000 resamples shows that the \textsc{Adversarial}, \textsc{Mask}, and \textsc{Mask + Adversarial} models outperform the baseline on the clean test set for both micro- and macro-averaged $F_1$ scores ($p < 0.001$). Among robustness-oriented strategies, the \textsc{Adversarial} model outperforms the \textsc{Mask + Adversarial} model on the clean test set, which in turn outperforms the \textsc{Mask} model ($p < 0.001$).

With respect to additional evaluation metrics, the \textsc{Mask}, \textsc{Adversarial}, and \textsc{Mask + Adversarial} models all achieve higher recall than the baseline across micro, macro, and core-PT evaluations. The \textsc{Mask + Adversarial} model achieves the highest AUPRC under both micro-averaging (0.741) and macro-averaging (0.721), whereas the \textsc{Adversarial} model yields the lowest expected calibration error across all metrics ($\ell_1$ = 0.154; $\ell_2$ = 0.176; $\ell_{\max}$ = 0.318).

\begin{figure*}[ht]
    \centering
    \includegraphics[width=\textwidth]{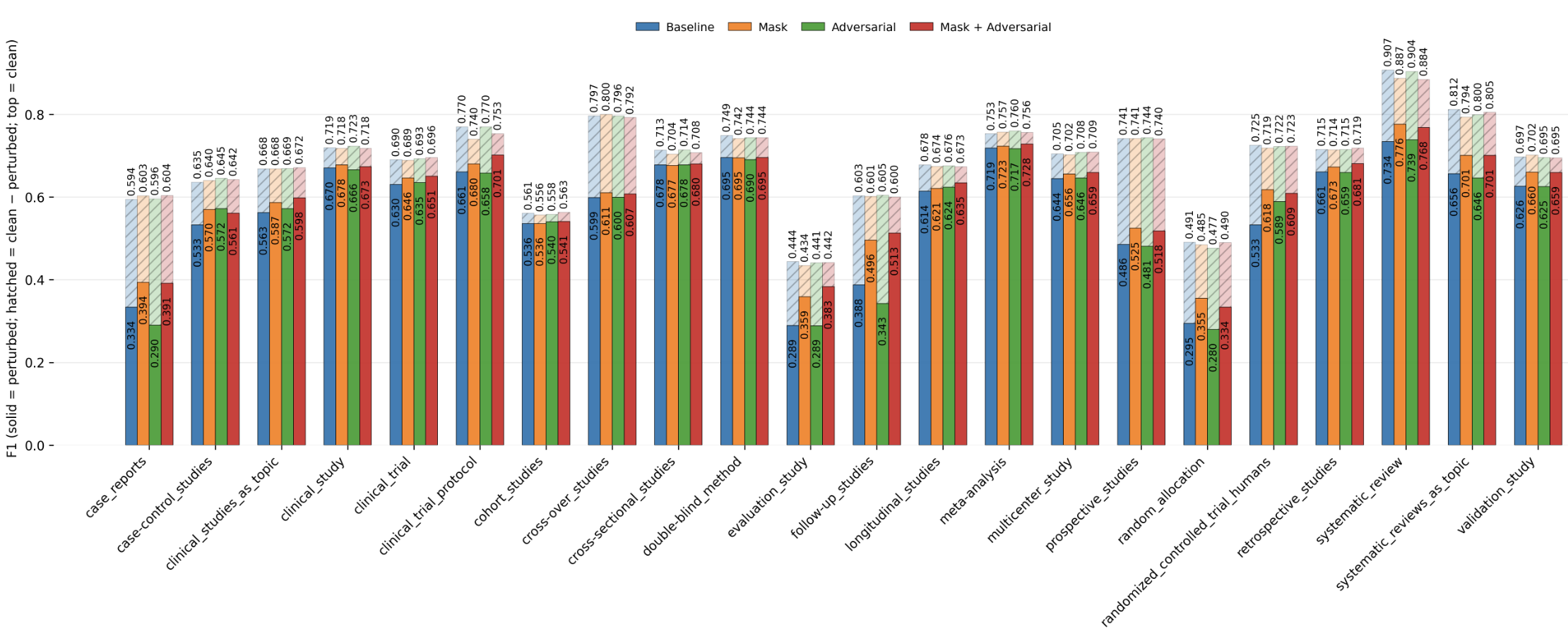}
    \caption{Label-wise performance degradation ($\Delta$F$_1$) under extreme semantic perturbation for core PTs.}
    \label{core}
\end{figure*}

\subsection{Robustness of Core Publication Types}
Semantic perturbations affect PT predictions to varying degrees. Figure \ref{core} presents label-wise micro-$F_1$ scores for the core PTs on the clean test set and the corresponding performance degradation ($\Delta F_1$) under extreme perturbation for the baseline, \textsc{Mask}, \textsc{Adversarial}, and \textsc{Mask + Adversarial} models.

Several PTs exhibit relatively high robustness across models. For example, \textit{Cohort Studies} shows minimal degradation, with the smallest drop ($\Delta F_1 = 0.02$) achieved by the \textsc{Adversarial} model. \textit{Meta-analysis} also remains comparatively stable under perturbation, with the smallest degradation observed for the \textsc{Mask + Adversarial} model ($\Delta F_1 = 0.028$). \textit{Cross-sectional Studies} similarly exhibits small performance drops across models. In contrast, \textit{Case Reports} and \textit{Prospective Studies} exhibit larger performance degradation across all models.

Certain PTs benefit more substantially from robustness-oriented training. For example, \textit{Randomized Controlled Trial Humans} shows a large performance drop for the baseline model under extreme perturbation ($\Delta F_1 = 0.192$), which is substantially reduced by masked-entity training ($\Delta F_1 = 0.101$) and the \textsc{Mask + Adversarial} strategy ($\Delta F_1 = 0.114$). A similar trend is observed for \textit{Case-Control Studies}, where the baseline model shows a larger performance drop ($\Delta F_1 = 0.102$) than the \textsc{Mask} model ($\Delta F_1 = 0.070$).

\section{Discussion}

\subsection{Sensitivity of PT Classifiers to Controlled Semantic Shifts and the Effects of Robustness-Oriented Training}

Our results show that the baseline model relies, to some extent, on domain-specific lexical cues, making it vulnerable to semantic shifts that preserve methodological meaning. This sensitivity becomes increasingly pronounced as perturbation severity increases, particularly under concept substitution, where topical content and entity semantics are altered beyond surface lexical variation. Robustness-oriented training strategies mitigate this sensitivity, with their benefits becoming more apparent under stronger perturbations.

Masked-entity training improves robustness by limiting the model's access to surface-form biomedical entities, resulting in more stable predictions. In contrast, domain-adversarial training regularizes learned representations to be invariant to domain-related signals and contributes most strongly to improved probabilistic calibration and clean-data performance, yielding more reliable confidence estimates under semantic shift. The \textsc{Mask + Adversarial} model leverages these complementary effects, providing the most consistent robustness improvements across perturbation settings while preserving strong in-domain performance.

These findings indicate that the commonly observed trade-off between robustness and in-domain accuracy can be mitigated when robustness objectives are designed to selectively suppress task-irrelevant features. By masking topical, non–PT-defining biomedical entities at the input level and combining this strategy with domain-adversarial regularization, our approach reduces spurious correlations at multiple stages of model learning. This observation aligns with recent work demonstrating that restricting adversarial perturbations to non-salient features can simultaneously improve performance on clean inputs and robustness under perturbation \cite{redgrave2025salient}.

\begin{table}[t]
\caption{Performance of masked-entity training with different masking ratios (30\%, 50\%, and 100\%) under clean and perturbed settings (100\% concept substitution + EDA). 95\% CIs via bootstrap ($n=1{,}000$).}
\centering

\scriptsize
\setlength{\tabcolsep}{2pt}
\renewcommand{\arraystretch}{1.05}

\begin{tabular}{llccc|ccc}
\toprule
& & \multicolumn{3}{c}{\textbf{Clean}} & \multicolumn{3}{c}{\textbf{Perturbed}} \\
\cmidrule(lr){3-5} \cmidrule(lr){6-8}
\textbf{Mask} & \textbf{Metric} 
& Micro & Macro & Core 
& Micro & Macro & Core \\
\midrule

\multirow{3}{*}{30\%}
& $F_1$
& \shortstack{\tiny \textbf{0.671}\\[-2pt]{\tiny [0.668--0.673]}}
& \shortstack{\tiny \textbf{0.678}\\[-2pt]{\tiny [0.671--0.684]}}
& \shortstack{\tiny \textbf{0.688}\\[-2pt]{\tiny [0.683--0.692]}}
& \shortstack{\tiny 0.600\\[-2pt]{\tiny [0.597--0.603]}}
& \shortstack{\tiny \textbf{0.604}\\[-2pt]{\tiny [0.596--0.613]}}
& \shortstack{\tiny 0.599\\[-2pt]{\tiny [0.595--0.604]}} \\

& Precision
& \shortstack{\tiny0.657\\[-2pt]{\tiny [0.654--0.661]}}
& \shortstack{\tiny0.688\\[-2pt]{\tiny [0.678--0.697]}}
& \shortstack{\tiny0.689\\[-2pt]{\tiny [0.684--0.694]}}
& \shortstack{\tiny0.645\\[-2pt]{\tiny [0.641--0.648]}}
& \shortstack{\tiny0.689\\[-2pt]{\tiny [0.677--0.700]}}
& \shortstack{\tiny0.669\\[-2pt]{\tiny [0.664--0.674]}} \\

& Recall
& \shortstack{\tiny0.684\\[-2pt]{\tiny [0.681--0.688]}}
& \shortstack{\tiny0.684\\[-2pt]{\tiny [0.678--0.690]}}
& \shortstack{\tiny0.697\\[-2pt]{\tiny [0.692--0.702]}}
& \shortstack{\tiny0.562\\[-2pt]{\tiny [0.558--0.565]}}
& \shortstack{\tiny0.565\\[-2pt]{\tiny [0.556--0.573]}}
& \shortstack{\tiny0.571\\[-2pt]{\tiny [0.566--0.577]}} \\

\midrule

\multirow{3}{*}{50\%}
& $F_1$
& \shortstack{\tiny0.668\\[-2pt]{\tiny [0.665--0.671]}}
& \shortstack{\tiny0.675\\[-2pt]{\tiny [0.667--0.681]}}
& \shortstack{\tiny0.685\\[-2pt]{\tiny [0.681--0.689]}}
& \shortstack{\tiny\textbf{0.601}\\[-2pt]{\tiny [0.598--0.605]}}
& \shortstack{\tiny\textbf{0.604}\\[-2pt]{\tiny [0.595--0.612]}}
& \shortstack{\tiny\textbf{0.601}\\[-2pt]{\tiny [0.597--0.606]}} \\

& Precision
& \shortstack{\tiny0.656\\[-2pt]{\tiny [0.653--0.659]}}
& \shortstack{\tiny0.689\\[-2pt]{\tiny [0.680--0.698]}}
& \shortstack{\tiny0.687\\[-2pt]{\tiny [0.683--0.693]}}
& \shortstack{\tiny0.647\\[-2pt]{\tiny [0.644--0.651]}}
& \shortstack{\tiny0.685\\[-2pt]{\tiny [0.671--0.696]}}
& \shortstack{\tiny0.666\\[-2pt]{\tiny [0.661--0.671]}} \\

& Recall
& \shortstack{\tiny0.680\\[-2pt]{\tiny [0.677--0.683]}}
& \shortstack{\tiny0.681\\[-2pt]{\tiny [0.674--0.686]}}
& \shortstack{\tiny0.694\\[-2pt]{\tiny [0.689--0.699]}}
& \shortstack{\tiny0.562\\[-2pt]{\tiny [0.558--0.565]}}
& \shortstack{\tiny0.567\\[-2pt]{\tiny [0.559--0.576]}}
& \shortstack{\tiny0.576\\[-2pt]{\tiny [0.571--0.581]}} \\

\midrule

\multirow{3}{*}{100\%}
& $F_1$
& \shortstack{\tiny0.661\\[-2pt]{\tiny [0.658--0.664]}}
& \shortstack{\tiny0.654\\[-2pt]{\tiny [0.646--0.662]}}
& \shortstack{\tiny0.680\\[-2pt]{\tiny [0.675--0.684]}}
& \shortstack{\tiny0.591\\[-2pt]{\tiny [0.588--0.594]}}
& \shortstack{\tiny0.586\\[-2pt]{\tiny [0.580--0.593]}}
& \shortstack{\tiny0.595\\[-2pt]{\tiny [0.590--0.600]}} \\

& Precision
& \shortstack{\tiny0.648\\[-2pt]{\tiny [0.645--0.651]}}
& \shortstack{\tiny0.683\\[-2pt]{\tiny [0.671--0.697]}}
& \shortstack{\tiny0.684\\[-2pt]{\tiny [0.679--0.690]}}
& \shortstack{\tiny0.644\\[-2pt]{\tiny [0.640--0.647]}}
& \shortstack{\tiny0.680\\[-2pt]{\tiny [0.667--0.688]}}
& \shortstack{\tiny0.674\\[-2pt]{\tiny [0.669--0.680]}} \\

& Recall
& \shortstack{\tiny0.674\\[-2pt]{\tiny [0.671--0.678]}}
& \shortstack{\tiny0.651\\[-2pt]{\tiny [0.644--0.658]}}
& \shortstack{\tiny0.687\\[-2pt]{\tiny [0.681--0.692]}}
& \shortstack{\tiny0.547\\[-2pt]{\tiny [0.543--0.551]}}
& \shortstack{\tiny0.542\\[-2pt]{\tiny [0.535--0.548]}}
& \shortstack{\tiny0.558\\[-2pt]{\tiny [0.553--0.564]}} \\

\bottomrule
\end{tabular}
\label{mask}
\end{table}

\subsection{Impact of Masking Strength on the Robustness--Accuracy Trade-off}

Our analysis shows that masking strength controls the trade-off between robustness and in-domain accuracy. As shown in Table~\ref{mask}, masking 30\% achieves the highest micro-$F_1$ on the clean test set, while performance degrades as masking strength increases. In contrast, under perturbation, 50\% masking yields the best overall performance across micro-, macro-, and core $F_1$, indicating improved robustness. These results suggest that while masking suppresses spurious correlations, overly aggressive masking can remove informative, label-relevant cues. This pattern is consistent with prior work showing that both robust and non-robust features can contribute useful signals for standard classification \cite{ilyas2019adversarial}. Masking 50\% of training instances provides the best balance between robustness and in-domain performance.

\subsection{Effect of Explicit Methodological Signals on Robustness}
\begin{figure}[t]
    \centering
    \includegraphics[width=\columnwidth]{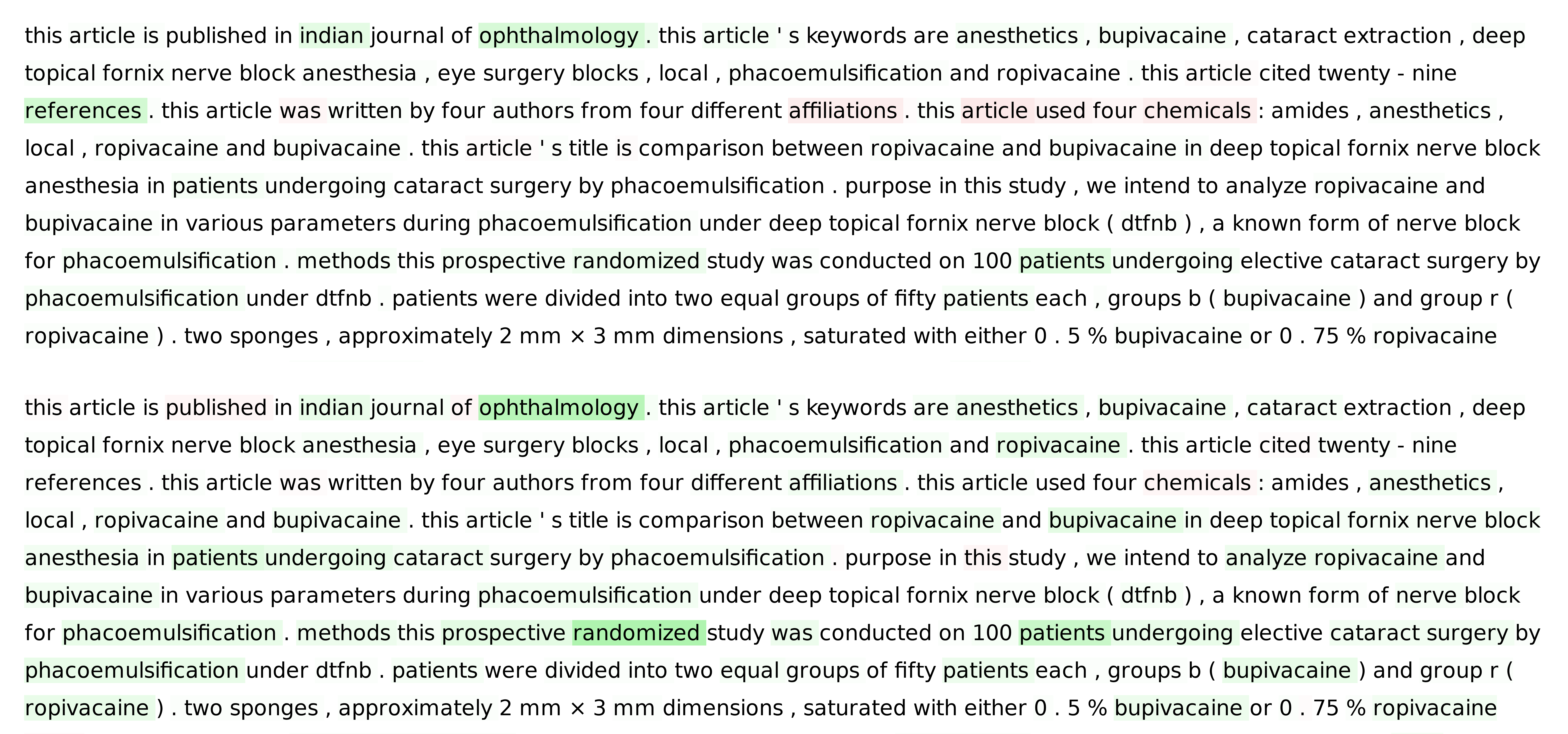}
    \caption{Saliency visualization illustrating a missed prediction of \textit{Randomized Controlled Trial Humans} by the baseline model (top; predicted probability = 0.452), contrasted with a correct prediction under \textsc{Adversarial} training (bottom; predicted probability = 0.696), where the adversarial model shifts attention toward methodological terms (PMID: 30127137).}
    \label{vis-1}
\end{figure}

To better understand how robustness-oriented training strategies influence model behavior, we conduct a qualitative analysis using token-level saliency maps. Figure \ref{vis-1} revisits the motivating example introduced earlier, in which the baseline model fails to predict the correct label \textit{Randomized Controlled Trial Humans}. Under \textsc{Adversarial} training, this error is corrected as the model reallocates attention toward methodological cues, prioritizing stable study-design descriptors such as ``prospective'' and ``randomized,'' which support the correct classification.

PT-defining terms may also be misinterpreted when they appear in strong topical contexts, leading to false-positive predictions. As shown in Figure \ref{vis-4}, the baseline model incorrectly associates the phrase ``outpatient follow-up of kidney transplant recipients,'' which conveys background information, with the \textit{Follow-Up Study} label. In contrast, the \textsc{Mask + Adversarial} model becomes more selective, down-weighting non–PT-specific cues (e.g., ``kidney,'' ``transplant,'' ``patients''). 

In both cases, explicit methodological cues relevant to study design are present in the input, and robustness-oriented training successfully reweights attention toward these signals, thereby correcting baseline misclassifications. To quantitatively assess whether the presence of explicit methodological terms affects the likelihood of error correction, we compare correction rates when PT-defining label-name terms are present in the input versus absent. Using Fisher's exact test, we find that the \textsc{Mask + Adversarial} model is more likely to correct baseline misclassifications when explicit methodological cues are present ($p < 0.01$).

\subsection{Large-Scale Attribution Analysis of Methodological vs. Topical Signals}
To assess whether these qualitative patterns generalize beyond individual examples, we analyze token-level normalized attribution scores produced by the baseline and \textsc{Mask + Adversarial} models at scale. Our analysis focuses on two classes of features: (1) PT-defining label-name terms (e.g., occurrences of ``cohort study'' for the \textit{Cohort Studies} label) and (2) topical biomedical entities identified using MetaMap and belonging to the categories \textit{Chemicals and Drugs, Anatomy, Disorders, Genes and Molecular Sequences, Geographic Areas}, and \textit{Physiology}. Attribution differences between models are evaluated using paired Wilcoxon signed-rank tests. When explicit PT-defining terms are present in the input, the \textsc{Mask + Adversarial} model assigns significantly higher attribution to these methodological cues than the baseline model ($p < 0.05$). In contrast, attribution assigned to topical biomedical entities is significantly reduced ($p < 0.001$).

\begin{figure}[]
    \centering
    \includegraphics[width=\columnwidth]{figure2_stacked_saliency.png}
    \caption{Excerpt from a saliency visualization illustrating a false-positive prediction of \textit{Follow-Up Study} by the baseline model (top; predicted probability = 0.571), contrasted with a corrected prediction by the \textsc{Mask + Adversarial} model (bottom; predicted probability = 0.482) (PMID: 27906832).}
    \label{vis-4}
\end{figure}

\begin{figure}[]
    \centering
    \includegraphics[width=\columnwidth]{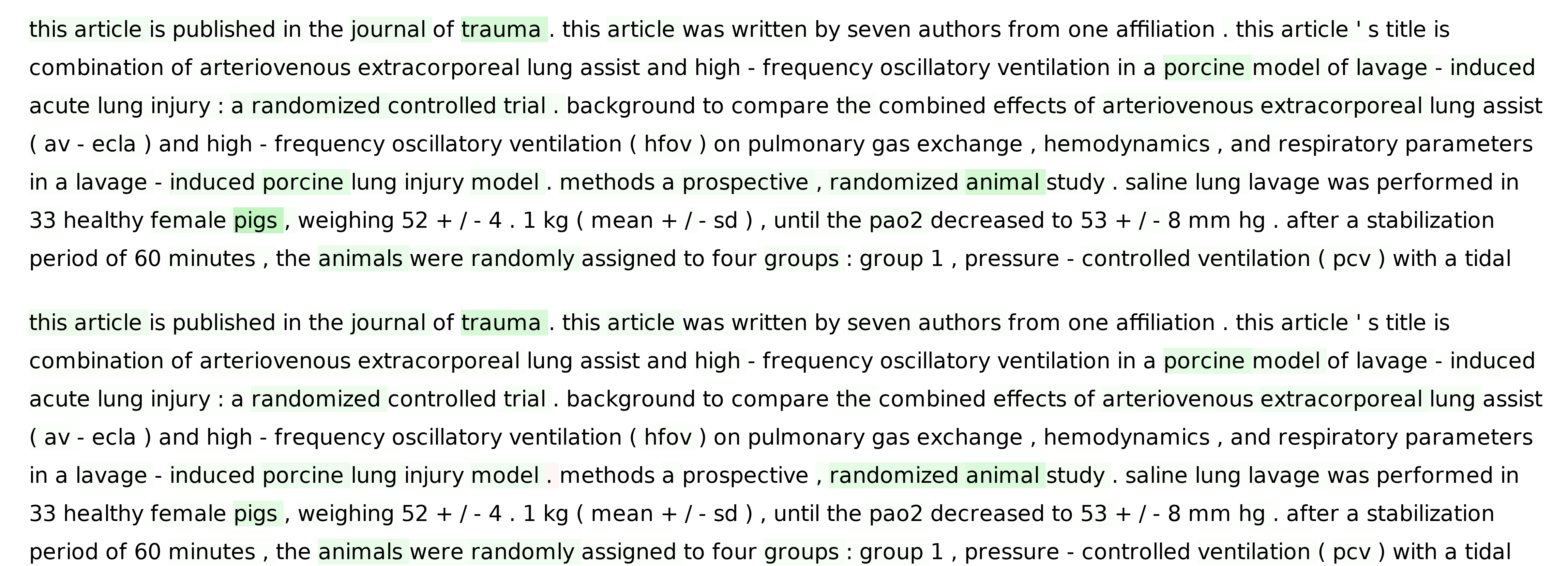}
    \caption{Excerpt from a saliency visualization illustrating a correct prediction of \textit{Veterinary Randomized Controlled Trials} by the baseline model (top; predicted probability = 0.559), but a false-negative prediction by the \textsc{Mask + Adversarial} model (bottom; predicted probability = 0.495) (PMID: 17297323).}
    \label{wrong}
\end{figure}

\subsection{Label-Specific Sensitivity and Implications for PT Prediction}
The label-wise robustness patterns observed in our analysis indicate that PT classifiers rely on different types of signals. PTs whose defining characteristics are expressed through explicit methodological or reporting cues tend to be more robust to semantic perturbations, as these signals remain stable even when topical content is altered. In contrast, PTs whose semantics are conveyed implicitly through narrative context are more sensitive to perturbation. In such cases, topical descriptions that are not inherently PT-specific may nonetheless carry legitimate design signals. For example, age–sex framing (e.g., ``a 45-year-old man'') strongly implies a single-patient narrative and therefore provides important evidence for \textit{Case Reports}. Suppressing such entity-driven cues can inadvertently remove task-relevant information. 

These label-dependent effects highlight that not all topical correlations are spurious. As illustrated in Figure \ref{wrong}, domain-adversarial training that down-weights animal-related cues by treating ``animal'' and its species-specific mentions as domain indicators rather than PT-defining features can lead to false-negative predictions.

Overall, robustness-oriented training is most effective when topical information primarily functions as a confounding signal rather than conveying task-relevant evidence. By limiting the model's access to such confounding cues, robustness objectives can encourage greater reliance on more stable, PT-defining patterns and improve robustness under semantic perturbations. In addition, these findings suggest that incorporating full-text inputs, particularly methods sections, may surface more explicit methodological structure that is often absent or weakly expressed in abstracts, thereby reducing reliance on confounded topical or narrative cues.

\subsection{Limitations and Future Directions}
While our robustness-oriented training strategies aim to reduce reliance on spurious topical shortcuts, certain forms of topical information (e.g., population characteristics) may legitimately correlate with specific PTs, and enforcing uniform domain invariance may suppress informative signals. In addition, our evaluation is conducted on a single encoder, SPECTER2-base, which limits conclusions about generalizability to other biomedical models such as PubMedBERT. To mitigate data sparsity in domain-adversarial training, we restrict the domain classifier to 53 MeSH-based labels using a 1\% frequency threshold; however, this heuristic may exclude rare but potentially informative domains.

\section{Conclusion}
In this work, we proposed a robustness evaluation framework for publication type and study design classification based on controlled perturbations that explicitly alter topical content while preserving methodological meaning. Using this framework, we showed that selectively masking non-task-defining features, when combined with representation-level domain-adversarial training, yields a more balanced and practically useful trade-off between robustness and in-domain accuracy.

Our analysis further revealed substantial variation in the sensitivity of core PTs to topical shifts, with the effectiveness of robustness-oriented training strategies differing accordingly. These findings suggest that PT-defining features range from explicit methodological cues to more implicit, context-dependent signals, with the latter remaining more vulnerable to perturbation across models. 

Beyond PT classification, the proposed perturbation framework provides a general diagnostic perspective for analyzing feature reliance in biomedical NLP systems. Similar methodology–topic disentanglement challenges arise in tasks such as reporting guideline adherence assessment \cite{jiang2025spirit} and evidence screening for systematic reviews, where distinguishing methodological relevance from topical content is critical. Applying controlled perturbation analyses in these settings may help identify spurious shortcuts, improve model reliability, and support more trustworthy evidence synthesis pipelines.

\section*{Acknowledgment}
This work was supported by the National Library of Medicine of the National Institutes of Health under the award number R01LM14292. The content is solely the responsibility of the authors and does not necessarily represent the official views of the National Institutes of Health. The funder had no role in considering the study design or in the collection, analysis, interpretation of data, writing of the report, or decision to submit the article for publication.

\bibliographystyle{IEEEtran}
\bibliography{ref}

\end{document}